\documentclass[conference]{IEEEtran}
\IEEEoverridecommandlockouts
\usepackage{amsmath,amssymb,amsfonts}
\usepackage{algorithmic}
\usepackage{graphicx}
\usepackage{textcomp}
\usepackage{xcolor}
\usepackage{multirow}
\usepackage[usestackEOL]{stackengine}
\usepackage{caption}
\usepackage{subcaption}
\usepackage{cite}
\usepackage[english]{babel}
\usepackage[autostyle, english=american]{csquotes}
\MakeOuterQuote{"}

\def\BibTeX{{\rm B\kern-.05em{\sc i\kern-.025em b}\kern-.08em
    T\kern-.1667em\lower.7ex\hbox{E}\kern-.125emX}}

\usepackage{amsfonts}
\usepackage{booktabs}
\usepackage{siunitx}
\usepackage{lipsum}
\usepackage{float}
\usepackage{color,soul} 

\begin{document}

\title{An Explainable and Conformal AI Model to Detect Temporomandibular Joint Involvement in Children Suffering from Juvenile Idiopathic Arthritis}

\author{
    Lena Todnem Bach Christensen\textsuperscript{*}, Dikte Straadt\textsuperscript{*}, Stratos Vassis\textsuperscript{\dag}, Christian Marius Lillelund\textsuperscript{*}\textsuperscript{\ddag},\\ Peter Bangsgaard Stoustrup\textsuperscript{\dag}, Ruben Pauwels\textsuperscript{\dag}, Thomas Klit Pedersen\textsuperscript{\dag} and Christian Fischer Pedersen\textsuperscript{*}\\
    Email: \{lena.todnem,dikte.straadt\}@gmail.com \{cl,cfp\}@ece.au.dk \\
    \{stratos.vassis,pstoustrup,ruben.pauwels,thomas.klit\}@dent.au.dk \\
    \textsuperscript{*}Department of Electrical and Computer Engineering, Aarhus University, Denmark \\
    \textsuperscript{\dag}Department of Dentistry and Oral Health, Aarhus University, Denmark \\
}

\maketitle
\begingroup\renewcommand\thefootnote{\ddag}
\footnotetext{Corresponding author: Christian Marius Lillelund (cl@ece.au.dk)}
\endgroup

\begin{abstract}
Juvenile idiopathic arthritis (JIA) is the most common rheumatic disease during childhood and adolescence. The temporomandibular joints (TMJ) are among the most frequently affected joints in patients with JIA, and mandibular growth is especially vulnerable to arthritic changes of the TMJ in children. A clinical examination is the most cost-effective method to diagnose TMJ involvement, but clinicians find it difficult to interpret and inaccurate when used only on clinical examinations. This study implemented an explainable artificial intelligence (AI) model that can help clinicians assess TMJ involvement. The classification model was trained using Random Forest on 6154 clinical examinations of 1035 pediatric patients (67\% female, 33\% male) and evaluated on its ability to correctly classify TMJ involvement or not on a separate test set. Most notably, the results show that the model can classify patients within two years of their first examination as having TMJ involvement with a precision of 0.86 and a sensitivity of 0.7. The results show promise for an AI model in the assessment of TMJ involvement in children and as a decision support tool.
\end{abstract}


\section{Introduction}
It is challenging for clinicians to diagnose early involvement of the temporomandibular joints (TMJ) in children with juvenile idiopathic arthritis (JIA), as the disease is often asymptomatic \cite{b1}. Inadequate management of potential involvement of the temporomandibular joints may lead to asymmetrical jaw growth, skeletal deformities, restricted maximal incisor opening, pain or discomfort in the jaw joint area and the surrounding muscles. Furthermore, a child may experience airway constriction and sleep-disordered breathing, which can disrupt their sleeping pattern and worsen their quality of life \cite{b1, b2, stoustrup_management_2023}. In this context, TMJ involvement means findings believed to be caused by previous or current TMJ arthritis. There are several modalities to diagnose TMJ arthritis, e.g., clinical (orthodontic) examinations, computed tomography (CT) scans, contrast-enhanced magnetic resonance imaging (MRI), and ultrasound \cite{b4}. Clinical examinations are the most cost-effective modality to assess TMJ involvement of patients with JIA \cite{b2}, however, no single parameter of the clinical examination can be suggested as a predictor of TMJ involvement in patients with JIA \cite{b2}, and the sensitivity and specificity of detecting TMJ involvement using clinical examinations are low compared to CT and MRI \cite{b2}, which on the other hand are costly and time-consuming \cite{b2}.

Müller et al. attempted early diagnosis of TMJ involvement using clinical examinations, ultrasound and MRI \cite{muller_early_2009}. Based on 33 patients with a diagnosis of JIA, they found that a clinical examination had a sensitivity of 0.47 (95\% CI = 0.25-0.71) for patients and 0.66 (95\% CI = 0.47-0.81) for joints. The agreement between clinical and MRI diagnoses was observed in 16 out of 30 patients (53\%) and 34 out of 60 joints (57\%). In particular, all 15 TMJs with condylar deformities detected by MRI were accurately diagnosed as pathological by clinical examination, but only 7 out of 18 with TMJ involvement (39\%) received a correct diagnosis. This motivates the use of additional decision support if clinical examinations are to be used as an indicator of early TMJ involvement. Previous work have mainly focused on the role of clinical examination in diagnosing TMJ arthritis (i.e., the presence of active TMJ arthritis), but to the best of our knowledge, no prior work has proposed a model that can identify TMJ involvement based on clinical examinations alone. In a recent consensus article on the management of the orofacial manifestation of JIA, clinical examination was recommended as a routine screening tool \cite{stoustrup_management_2023}.

In this work, we propose an explainable AI model to assess TMJ involvement in children under the age of 16 who suffer from JIA. The model is based exclusively on clinical examinations carried out in accordance with consensus-based standards for the specific patient group. The objective is to aid in early diagnosis by providing clinicians with a decision support tool that can inform them of significant clinical findings or symptoms that a patient may experience, that could indicate the presence of TMJ involvement. Early diagnosis and management can prevent some of the physical and psychological consequences that come with JIA, and lead to an anticipated better outcome for the child and their family.

\begin{figure*}
    \centering
    \includegraphics[width=0.75\linewidth]{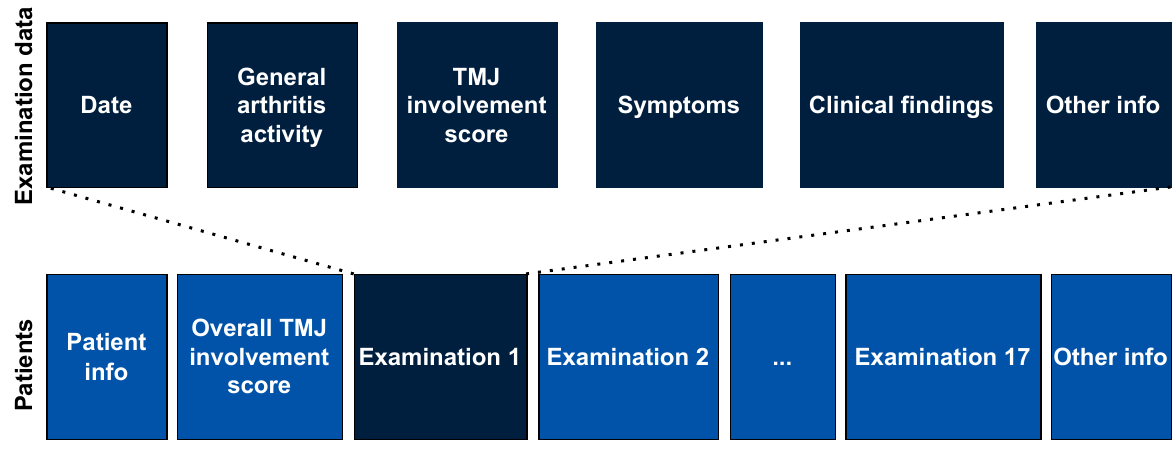}
    \caption{Each examination of a JIA patient includes a screening, where the patient's symptoms, drug usage and other clinical findings are recorded. The dataset contains historical examinations of 1035 JIA patients spanning a 25-year period with 6154 longitudinal records in total.}
    \label{fig:data_structure}
\end{figure*}

\section{Methodology}

\subsection{Clinical data}
The dataset contains 6154 longitudinal records of 1035 pediatric patients ($<$16 years old) over a 25-year period, recorded and provided by the Department of Dentistry and Oral Health, Aarhus University. Each patient underwent 2-17 clinical examinations, where the patient's symptoms, drug usage, general arthritis level, TMJ involvement and clinical findings were recorded (see Fig. \ref{fig:data_structure}). The clinical variables examined ($d=95$) at each visit are in agreement with evidence-based standards on orofacial examination in JIA \cite{Stoustrup2017ClinicalOE}. Among the cohort of patients, 690 are girls and 345 are boys, reflecting the gender distribution commonly observed in patients with JIA within an open population \cite{b6}. All patients were diagnosed with JIA prior to their first examination. However, not all patients initially showed signs of TMJ involvement; some developed it over time, while others were diagnosed with TMJ involvement after the examination was completed. The involvement of the TMJ was confirmed or refuted by an X-ray or MRI. Table \ref{tab:significant} shows all recorded clinical features ($d=95$) and a subset chosen by experts in the field ($d=26$). We provide a complete feature description table in the Supplementary file.

\subsection{Preprocessing and feature engineering}
We encode all categorical features using label encoding and utilize entity embeddings with an embedding vector dimensionality of one to encode nominal features. The two continuous features, \textit{openmm} and \textit{protrusionmm}, increase with age of the patient, so we apply a transformation to indicate the deviation of the patient from the gender and age specific averages. The \textit{drug} feature indicates what kind of medication being provided to the patient and includes 55 distinct combinations, so we split the feature into five subgroups of medication to reduce the number of features. Three general classes of medications are commonly prescribed to treat arthritis; they include non-steroidal anti-inflammatory drugs (NSAIDs), corticosteroids, and disease-modifying antirheumatic drugs (DMARDs). DMARDs can be further categorized into conventional DMARDs and biological DMARDs. By implementing this categorization, the feature space for the \textit{drug} feature was reduced from 55 to just five. In addition, clinical features extracted from the left and right side of the face have a strong correlation, so all side-specific features were merged into a single feature taking a common value or the highest value of the left and right feature. Lastly, we apply a $z$-score data normalization with zero mean and one standard deviation to all features before training.

\subsection{Classification analysis}
We propose a binary classification method to predict the involvement of TMJ in children with JIA. We adopt the acclaimed Random Forest machine learning algorithm \cite{b7}, since it is a flexible and easy-to-use machine learning algorithm, which has shown excellence on similar classification problems in the past. After preprocessing, we adopt an 80-10-10 data split; 80\% is used for training, 10\% is used for calibration, and 10\% is used for testing. We report the classifier's precision, sensitivity, and macro-averaged F1 score (F1\textsubscript{m}) on the test set (see Tab. \ref{tab:model_results}). The F1\textsubscript{m} is defined as: 

\begin{equation}\label{eq:f1}
F1 = 2 \cdot \frac{Precision \cdot Sensitivity}{Precision + Sensitivity}\text{,}
\end{equation}
\begin{equation}\label{eq:f1_macro}
F1_{m} = \frac{F1_{class_1} + F1_{class_2} + ... + F1_{class_i}}{N}\text{.}
\end{equation}

\subsection{Uncertainty estimation}
To quantify the uncertainties of the predictive model, we use the Model Agnostic Prediction Interval Estimator (MAPIE) library to train a conformal classifier \cite{b11}. The classifier wraps the original classifier and produces conformal prediction sets that have a guaranteed marginal coverage rate. Thus, the model output is then a set of classes instead of a single class. We set $\alpha=0.1$, indicating that we can expect 90\% of the prediction regions to cover the true outcome. However, the guarantee is only an average for the samples from the distribution of the calibration data. In practice, we use the Rescaled Adjusted Partial Sums (RAPS) method with a penalty term to reduce the size of the prediction sets.

\noindent
\begin{table}[t]
\centering
\caption{All clinical features ($d=95$), Expert features ($d=26$). See the Supplementary file for a full description.}
\label{tab:significant}
\begin{tabular}{p{8.2cm}}
\toprule
\textbf{All clinical features} ($d=95$) \\
\midrule
abrasion, aplasia, asybasis, asymenton, asyoccl, asypupilline, asyupmid, asymmetrymasseterright, asymmetrymasseterleft, backbending, bruxism, chewingfunction, clickclosingright, clickclosingleft, clicklateroleftright, clicklateroleftleft, clicklaterorightright, clicklaterorightleft, clickopeningright, clickopeningleft, clickprotrusionright, clickprotrusionleft, crepitationleft, crepitationright, deepbite, drug, dualbite, forwardbending, headache, hypermobilityleft, hypermobilityright, incisaloverjet, involvementstatus, krepitationleft, krepitationright, laterpalpleft, laterpalpright, laterotrusionleftmm, laterotrusionrightmm, lockleft, lockright, lips, lowerface, masseterleft, masseterright, micrognathism, morningstiffness, muscularpainleft, muscularpainright, neckpain, neckpalpation, neckstiffness, opening, openingfunction, openingmm, overbite, overjet, painleft, painmoveright, painmoveleft, ptext, ptint, profile, protrusion, protrusionmm, respiration, rotationleft, rotationright, sagittalrelationleft, sagittalrelationright, spacerelationship, sternoleft, sternoright, swollenjointright, swollenjointleft, swollenleft, swollenright, swollenjointright, swollenjointleft, swollenleft, swollenright, swollenright, temporalisleft, temporalisright, tempsenleft, tempsenright, tongue, tractionleft, tractionright, transversal, translationleft, translationright, dualbite, laterpalpright, laterpalpleft, postpalpright, postpalpleft \\\\
\toprule
\textbf{Expert-chosen clinical features} ($d=26$) \\
\midrule
asybasis, asyoccl, asypupilline, chewingfunction, deepbite, drug, krepitationleft, krepitationright, laterotrusionleftmm, laterotrusionrightmm, laterpalpleft, laterpalpright, lowerface, openbite, opening, openingmm, overbite, overjet, painmoveleft, painmoveright, profile, protrusion, protrusionmm, retrognathism, translationleft, translationright \\
\bottomrule
\end{tabular}
\end{table}

\subsection{Explainability}

Explainability helps the clinician judge whether the AI’s prediction is reasonable, emphasizing clinical features of highest importance and their relationship  with the predicted outcome. This can help turn AI from a black box into something that can be linked to current clinical knowledge. This also facilitates a comparative analysis of feature rankings across different feature spaces, assisting in the identification of overlapping features and their consistent significance across varied model configurations. We obtain explainability by computing Shapley Additive exPlanations (SHAP) values on the test set using the SHAP library \cite{b12}.

\section{Experiments and results}
Table \ref{tab:model_results} reports the predictive performance of the classifier. We adopt three sampling strategies (IID Data, Temporal Segmentation and Lagged Features) of the clinical dataset and five evaluation metrics for each class (precision, sensitivity, macro-averaged F1 score, coverage, and set size). This section will review the sampling strategies and the classifier's results.

\subsection{Strategy 1: IID Data}
\label{sec:iid_dataset}

This strategy assumes that all clinical examinations are IID (independent and identically distributed), i.e., the occurrence of TMJ involvement does not change over time. This approach neglects any temporal correlation between examinations and treats them all as independent. Using all features or only the expert features yields similar predictive performance with a macro F1 score of 0.87. However, this approach has poor practical application, as it involves examinations from the early stages of the disease, where involvement is less evident, to later stages of the disease, where involvement is clearly evident.

\subsection{Strategy 2: Temporal segmentation}
\label{sec:temp_segmented_dataset}

This strategy splits examinations in three intervals: 0-2, 2-5 and 5-15 years from a patient's first examination. Thus, the first model is trained on 2682 examinations recorded in the first two years, which can be used to determine early TMJ involvement in children when they have had their first or second examination at the clinic. We see a significant drop in predictive performance of about 12.5\% (from 0.8 to 0.7 sensitivity for the TMJ class), which is worse when training a model on examinations from the second to fifth year with a 27.5\% drop in sensitivity (from 0.8 to 0.58). Based on the present data, it is evidently easier to detect TMJ involvement in the early stage (0-2 years) of the disease than in the middle stage (2-5). After fives years, the clinical examination can be done more thoroughly, as the clinical signs are more evident, e.g., asymmetry starts to be visible at 4-5 years; this is related to patient maturity and progression of clinical signs.

\subsection{Strategy 3: Lagged Features}
\label{sec:lag_dataset}
This strategy converts the timeseries dataset to a supervised-learning problem by creating columns of lag samples that contain historical feature values. The first model is trained on 4983 samples with one lag feature for each feature, and the second model is trained on 3969 samples with two lag features. This naturally increases the number of features compared to the baseline (IID Data) strategy. Adding historical information increases the sensitivity slightly by 2.5\% (from 0.8 to 0.82) between the IID model using expert features only ($N=6006$, $d=26$) and the model with feature values from the previous observation ($N=4983$, $d=52$).

\begin{table}[!t]
\setlength{\tabcolsep}{3pt}
\centering
\caption{Performance metrics on the test set. $N$ is the number of samples in the dataset and $d$ is the number of features. The predicted classes, TMJ\textsubscript{0} and TMJ\textsubscript{1}, correspond to no and present TMJ involvement, respectively.}
\label{tab:model_results}
\resizebox{\columnwidth}{!}{%
\begin{tabular}{cccccccc}
\toprule
{\centering\Centerstack{Strategy}} &
{\centering\Centerstack{Dimensions}} & Class &
Precision $\uparrow$ & Sensitivity $\uparrow$ & F1\textsubscript{m} $\uparrow$ & Coverage $\uparrow$ & Set size $\downarrow$\\
\midrule
\multirow{3}{*}{\Centerstack{Sec.\\\ref{sec:iid_dataset}}}
& \Centerstack{$N=6006$\\ $d=95$} & \Centerstack{TMJ\textsubscript{0}\\TMJ\textsubscript{1}} & \Centerstack{0.88\\0.88} & \Centerstack{0.93\\0.81} & \Centerstack{0.8742} & \Centerstack{0.986\\0.899} & \Centerstack{1.154\\1.207} \\
\cmidrule(lr){3-8}
& \Centerstack{$N=6006$\\ $d=26$} & \Centerstack{TMJ\textsubscript{0}\\TMJ\textsubscript{1}} & \Centerstack{0.88\\0.88} & \Centerstack{0.93\\0.80} & \Centerstack{0.8706} & \Centerstack{0.986\\0.899} & \Centerstack{1.154\\1.207} \\
\cmidrule(lr){2-8}
\multirow{6}{*}{\Centerstack{Sec.\\\ref{sec:temp_segmented_dataset}}}
& \Centerstack{$N=2682$\\ $d=26$} & \Centerstack{TMJ\textsubscript{0}\\TMJ\textsubscript{1}} & \Centerstack{0.89\\0.86} & \Centerstack{0.95\\0.70} & \Centerstack{0.8455} & \Centerstack{0.969\\0.740} & \Centerstack{1.036\\1.078} \\
\cmidrule(lr){3-8}
& \Centerstack{$N=1430$\\ $d=26$} & \Centerstack{TMJ\textsubscript{0}\\TMJ\textsubscript{1}} & \Centerstack{0.76\\0.79} & \Centerstack{0.90\\0.58} & \Centerstack{0.7451} & \Centerstack{0.988\\0.737} & \Centerstack{1.233\\1.333} \\
\cmidrule(lr){3-8}
& \Centerstack{$N=1839$\\ $d=26$} &\Centerstack{TMJ\textsubscript{0}\\TMJ\textsubscript{1}} & \Centerstack{0.86\\0.78} & \Centerstack{0.75\\0.88} & \Centerstack{0.8144} & \Centerstack{0.967\\0.978} & \Centerstack{1.565\\1.326} \\
\cmidrule(lr){2-8}
\multirow{3}{*}{\Centerstack{Sec.\\\ref{sec:lag_dataset}}}
& \Centerstack{$N=4983$\\ $d=52$} & \Centerstack{TMJ\textsubscript{0}\\TMJ\textsubscript{1}} &
\Centerstack{0.88\\0.84} & \Centerstack{0.90\\0.82} & \Centerstack{0.8603} & \Centerstack{0.949\\0.906} & \Centerstack{1.233\\1.212} \\
\cmidrule(lr){3-8}
& \Centerstack{$N=3969$\\ $d=78$} & \Centerstack{TMJ\textsubscript{0}\\TMJ\textsubscript{1}} & \Centerstack{0.88\\0.84} & \Centerstack{0.89\\0.83} & \Centerstack{0.8578} & \Centerstack{0.961\\0.917} & \Centerstack{1.301\\1.310} \\
\bottomrule
\end{tabular}
}
\end{table}

\begin{figure*}
  \begin{subfigure}{0.33\textwidth}
    \includegraphics[width=\linewidth,trim={0 0 4cm 0},clip]{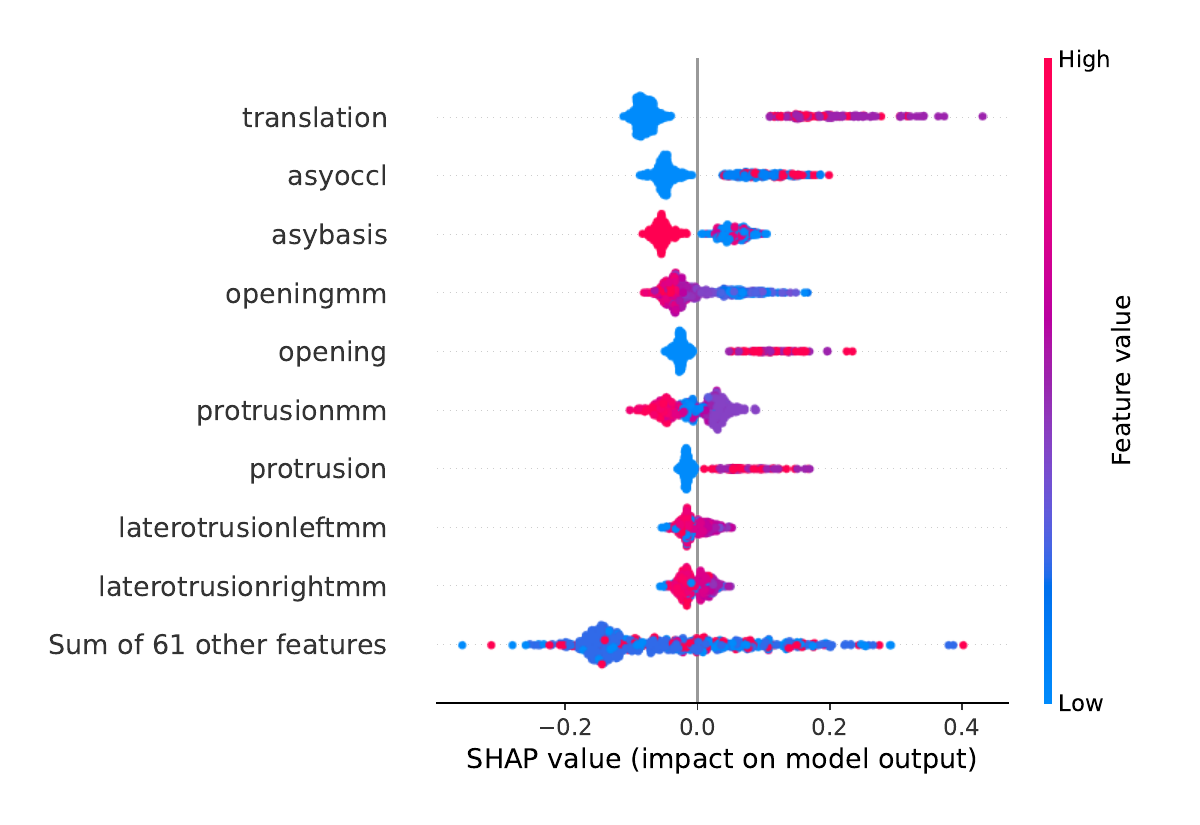}
    \caption{All samples together.}
    \label{fig:shap_plots_iid}
  \end{subfigure}%
  \begin{subfigure}{0.33\textwidth}
    \includegraphics[width=\linewidth,trim={0 0 4cm 0},clip]{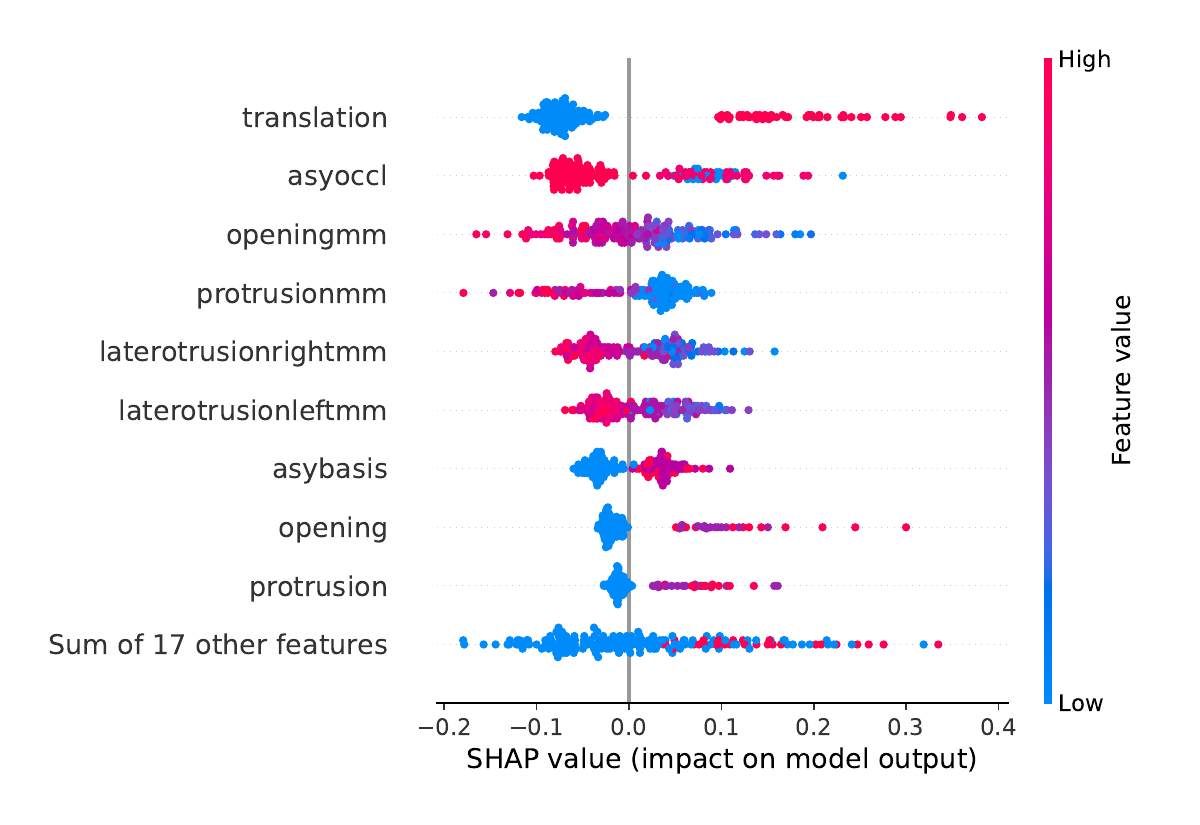}
    \caption{Between 5-15 years.}
    \label{fig:shap_plots_temporal}
  \end{subfigure}%
  \begin{subfigure}{0.33\textwidth}
    \includegraphics[width=\linewidth,trim={0 0 4cm 0},clip]{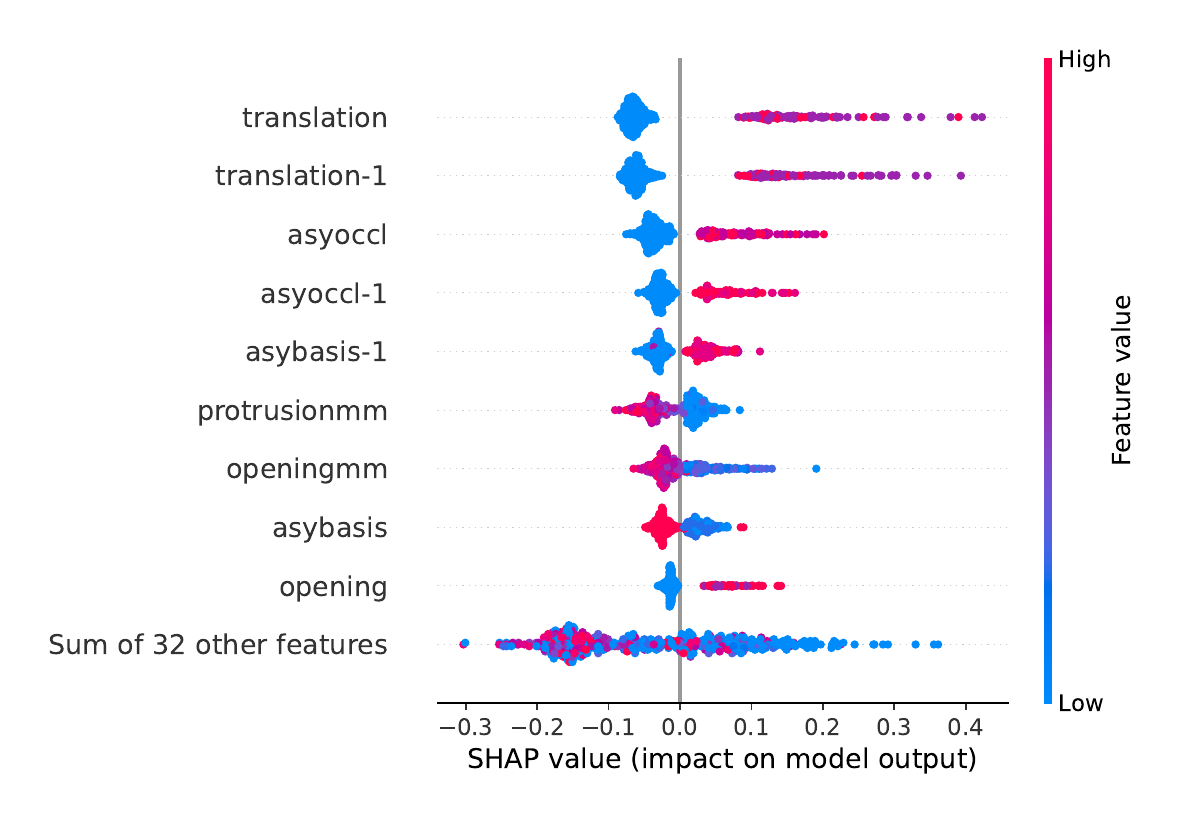}
    \caption{One previous observation.}
    \label{fig:shap_plots_lag}
  \end{subfigure}
  \caption{SHAP summary plots. The $y$-axis displays features, while the $x$-axis represents SHAP values. Features are ranked by importance, determined by the mean of their absolute SHAP values, with higher positions indicating greater significance. Each feature is represented by individual points, with a point corresponding to a single examination. The position on the $x$-axis signifies the impact of that feature on the model's output. The color of each point indicates whether the patient had a low (blue) or high (red) value of that specific feature. All SHAP values are computed on the test set.}
  \label{fig:shap_plots}
\end{figure*}

\section{Discussion}

The TMJ is among the most frequently affected joints in patients with JIA and mandibular growth is especially vulnerable to arthritic changes of the TMJ in children. The primary results of this study show that the proposed model with application of the "Temporal Segmentation" strategy (see Sec. \ref{sec:temp_segmented_dataset}) can classify patients early. That is, within a period of two years (i.e., from the first visitation and two years ahead) the model can classify patients with TMJ involvement with a precision of 0.86 and a sensitivity of 0.7. This means that out of all patients predicted as having TMJ involvement, 86\% had actual TMJ involvement, and 70\% of patients with TMJ involvement were correctly identified. This is based on a dataset with 2682 samples and 26 clinical features. Compared to \cite{muller_early_2009}, who attempted to detect TMJ involvement within 1 month of the MRI for 33 patients, they report a 0.47 sensitivity (95\% CI: 0.25-0.71) and a misdiagnosis rate of 61\%. Therefore, using our model offers a 49\% increase in sensitivity, although we note that our dataset is significantly bigger. The results show promise for an AI model in the assessment of TMJ involvement in children and as a decision support tool for clinicians. Such a tool has the potential to enhance timely diagnosis, which can improve the treatment outcome, and also assist clinicians in treatment planning.

\section{Conclusion}

The proposed method can be applied to current clinical examination results to help clinicians assess involvement of the temporomandibular joints in future patients, allowing for early and personalized follow-up.

\bibliographystyle{IEEEtran}
\bibliography{refs.bib}

\end{document}